\documentclass[10pt,twocolumn,letterpaper]{article}

\usepackage{cvpr}              

\usepackage[dvipsnames]{xcolor}

\definecolor{cvprblue}{rgb}{0.21,0.49,0.74}
\usepackage[pagebackref,breaklinks,colorlinks,citecolor=cvprblue]{hyperref}
\usepackage{graphicx}
\usepackage{amsmath}
\usepackage{amssymb}
\usepackage{booktabs}

\usepackage{algorithm}
\usepackage{algorithmic}
\usepackage{amsfonts}
\usepackage{xcolor}
\usepackage{color}
\usepackage{colortbl}
\usepackage{multirow}
\usepackage{subcaption}
\usepackage{newfloat}
\usepackage{listings}
\definecolor{cadmiumgreen}{rgb}{0.0, 0.42, 0.24}
\definecolor{ao}{rgb}{0.0, 0.5, 0.0}

\def\paperID{1378} 
\def\confName{CVPR}
\def\confYear{2024}

\title{Free Lunch for Gait Recognition: A Novel Relation Descriptor}

\author{Jilong Wang$^{1,2,4}$, Saihui Hou$^{3,4}$, Yan Huang$^{2}$, Chunshui Cao$^{4}$, Xu Liu$^{4,}$, \\
Yongzhen Huang$^{3,4}$, Tianzhu Zhang$^{1}$, Liang Wang$^{2}$\thanks{Corresponding Author} \\[2mm]
$^1$University of Science and Technology of China\\
$^2$Institute of Automation, Chinese Academy of Sciences\\
$^3$Beijing Normal University\\
$^4$WATRIX.AI\\
}

\begin{document}
\maketitle

\begin{abstract}
Gait recognition is to seek correct matches for query individuals by their unique walking patterns.
However, current methods focus solely on extracting individual-specific features, overlooking ``interpersonal" relationships.
In this paper, we propose a novel \textbf{Relation Descriptor} that captures not only individual features but also relations between test gaits and pre-selected gait anchors.
Specifically, we reinterpret classifier weights as gait anchors and compute similarity scores between test features and these anchors, which re-expresses individual gait features into a similarity relation distribution.
In essence, the relation descriptor offers a holistic perspective that leverages the collective knowledge stored within the classifier's weights, emphasizing meaningful patterns and enhancing robustness.
Despite its potential, relation descriptor poses dimensionality challenges since its dimension depends on the training set's identity count.
To address this, we propose Farthest gait-Anchor Selection to identify the most discriminative gait anchors and an Orthogonal Regularization Loss to increase diversity within gait anchors.
Compared to individual-specific features extracted from the backbone, our relation descriptor can boost the performance nearly without any extra costs.
%
%
We evaluate the effectiveness of our method on the popular GREW, Gait3D, OU-MVLP, CASIA-B, and CCPG, showing that our method consistently outperforms the baselines and achieves state-of-the-art performance.

\end{abstract}

\section{Introduction}
Gait recognition aims at identifying people at a long distance by their unique walking patterns \cite{survey}. 
As an identification task in vision, the essential goal of it is to learn the distinctive and invariant representations from the physical and behavioral human walking characteristics. 
With the boom of deep learning, gait recognition has achieved significant progress \cite{gaitgan,geinet,gaitpart,gaitset,opengait,wang2023causal}, yielding impressive results on public datasets. 
%
\begin{figure}[t]
     \centering
     \includegraphics[width=\linewidth]{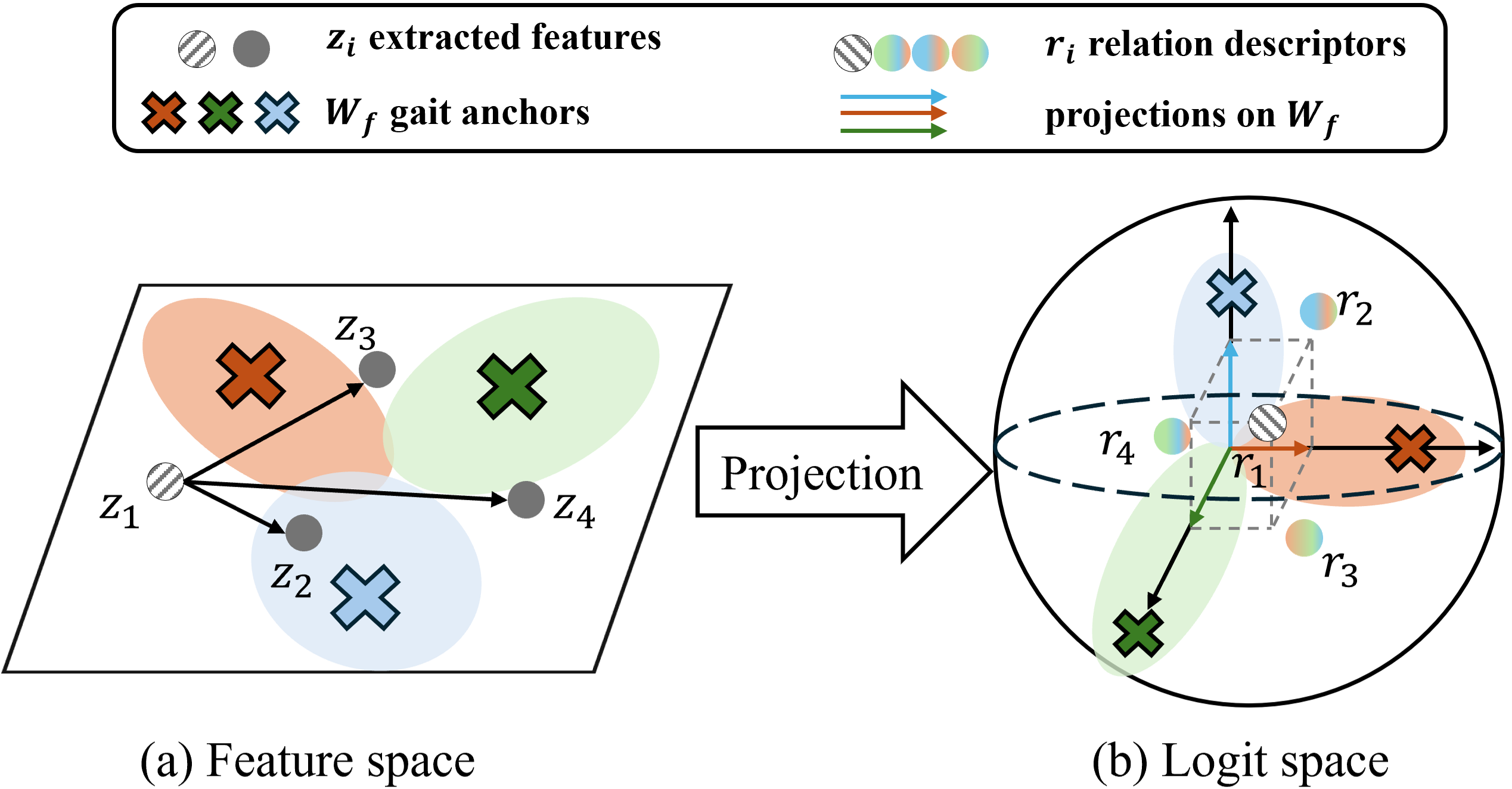}
     \caption{The comparison of identity-specific embeddings and relation-specific logits. (a) Conventional gait recognition utilizes the extracted feature for the final identification. (b) Gait is described by the similarity to fixed semantic directions by projecting gait features on well-trained gait anchor's vectors.}
     \label{fig:intro}
 \end{figure}


Reappraising an established pipeline of gait recognition \cite{GLN,GaitGL,opengait,dygait}, it typically involves a \textit{feature extractor} for obtaining unique gait features of individual walking sequences and usually utilizes a \textit{classifier} \cite{bnneck} for accurate identity classification. 
In common practice, the output embedding of the feature extractor is used for the test, while the classifier is usually discarded since testing and training identities are different.
However, the classifier usually occupies the majority of the network's weights, especially when the number of training classes is large, for example, 92.1\% and 75.1\% of total weights in GaitBase \cite{opengait} on GREW \cite{GREW} and OUMVLP \cite{OUMVLP}, respectively. 
Thus, a natural question arises: \textit{whether the well-trained weights in classifiers are really useless for inference?}

In this work, we draw wisdom from human beings: \textbf{human nature is the ensemble of social relations} \cite{jaggar1983feminist}.
%
This philosophical point of view provides a new perspective on gait recognition.
Gait features are sensitive to many covariant factors such as viewpoints and clothing, resulting in challenges of reducing intra-identity variations \cite{wang2023causal, gaitview}. Thus, capturing the accurate features under different conditions becomes difficult.
Intuitively, providing a relation descriptor within a whole group, such as ``tallest person among them", sometimes makes it easier to identify a person than just giving an inherent feature like ``7 feet tall".
%
Inspired by it, we assume that gait goes beyond just an aggregation of individual features, and it can also be expressed through the relationships with the gait features of others.
%
%
%
These relationships may involve aspects of similarity, dissimilarity, common traits, \etc, reflecting patterns and variations of gait within a population.
As shown in \Cref{fig:intro},  gait features can be described by the difference/similarity relationships with the several pre-selected people's \textbf{G}ait \textbf{A}nchors (GAs). 

To get a relation descriptor, a set of GAs should be first determined. 
A good set of GAs should contain various gait patterns. For instance, if we select all GAs with similar body shapes and postures, the relation descriptor loses its discriminative capacity since it fails to reflect distinct features of gait.
In our work, we innovatively find that the weights in the classifier are suitable for severing as GAs by reinterpreting it as \textit{the well-defined gait prototypes of different people in the training set}.
%
%
Hence, the projection of a gait embedding onto these prototypes can be used as a representation that describes their relationships.
Specifically, the normalized dot product of gait features and GAs generates a distribution of similarities, constructing a new \textbf{Relation Descriptor} (RD).
Essentially, \textit{RD offers a holistic perspective that leverages the collective knowledge stored within the classifier's weights.}
In principle, we find RD brings two benefits: 1) emphasizing meaningful features instead of noise and 2) enhancing robustness and generalizability.

However, directly employing RD poses two challenges, \ie, \textbf{dimension expansion} and \textbf{GAs overfitting}, as the number of GAs in the classifier depends on the count of training identities.
When numerous GAs are selected, such as 20000 and 5153 in GREW \cite{GREW} and OU-MVLP \cite{OUMVLP}, the dimension of RD can largely surpass that of the original embedding, resulting in increased storage costs and practical challenges in real-world applications. 
%
%
On the other hand, we find that too few identities lead to an overfitting problem that all GAs are highly related, reducing the variety of gait patterns and variations within GAs.
Therefore, balancing the quantity of GAs to avoid excessive expansion and removing the correlation between GAs are key considerations in applying RD. 
%

For the problem of \textbf{dimension expansion}, we find that not all identities in the training set are needed since there would be many similar identity prototypes, which results in redundant relationships.
Inspired by Archetypal Analysis \cite{AAface}, we assume that the most discriminative combination of GAs in the latent space should be the one with the largest spanning space, \emph{i.e.}, the convex combinations of the archetypes. 
Accordingly, we introduce a \textbf{F}arthest gait-\textbf{A}nchor \textbf{S}election (FAS) algorithm to select the most discriminative set of GAs. 
%
%
For the problem of \textbf{GAs overfitting}, we find the main cause is a target misalignment between cross-entropy loss and GAs. When the number of identities is fewer than the embedding dimension, the network could easily push the logits of the ground truth (GT) class to 1 and the rest to -1, measured by cosine similarity.
However, we desire that a sample only related to its own class weight, which means it should be orthogonal to other class weights.
Therefore, we propose an \textbf{O}rthogonal \textbf{R}egularization \textbf{L}oss (ORL) for better classifier training, encouraging the cosine similarity of a sample to its own class to be close to $1$ and the similarity to other classes to be close to $0$ instead of $-1$.
As a result, RD can better reflect the distinct characteristics of different individuals' gaits, even with a small set of GAs.

We rigorously evaluate our proposed approach on GREW \cite{GREW}, Gait3D \cite{Gait3d}, OU-MVLP \cite{OUMVLP}, CASIA-B \cite{CASIAB}, and CCPG \cite{ccpg}, consistently demonstrating its superiority over conventional baseline methods. To summarize, the contributions of our work can be outlined in three aspects:

(i) We propose a novel descriptor for gait recognition, capturing not only individual features but also relationships among well-trained gait anchors, which enhances recognition performance nearly without extra costs.

(ii) We address the challenges of dimension expansion and GAs overfitting by the Farthest gait-Anchor Selection algorithm and Orthogonal Regularization Loss, improving efficiency and discrimination.

(iii) We evaluate the effectiveness of our proposed method on five popular datasets, and the extensive experimental results demonstrate the superior performance of our approach, \eg, 5.5\% and 5.4\% absolute improvements on Gait3D and GREW in terms of rank-1 accuracy.

\section{Related Work}
\subsection{Gait Recognition}
Gait recognition identifies people by their unique representation of gait characteristics, which is easily affected by many covariant factors.
Previous works usually introduce extra information or design a better feature extractor to learn an invariant representation,
which can be roughly grouped into model-based \cite{modelbased1, modelbased2, modelbased3, modelbased4} and appearance-based \cite{GaitGL,gaitpart,gaitset,GEI,GLN,geinet} categories according to the type of input. 
%
%
%
Thanks to prior works' contributions, a typical gait recognition pipeline has been established. 
It primarily consists of four components: 1) a spatial-temporal feature extractor to get individual inherent features (CNN \cite{gaitset,opengait,gaitpart,GaitGL,dygait,dou2022metagait}, Transformer \cite{fan2023deepgait,wu2023gaitformer}, GCN \cite{teepe2021gaitgraph, fu2023gpgait}, \etc), 2) a temporal pooling module to aggregate sequence's features (MaxPooling \cite{gaitset}, MeanPooling \cite{geinet}, GeM \cite{GaitGL}, \etc), 3) a multi-scale module to obtain fine-grained information  (HPP \cite{gaitset}, attention \cite{dou2022metagait,dou2023gaitgci}, body parsing, \etc), 4) and metric learning loss functions to enhance features' discrimination (Triplet Loss \cite{triplet}, Cross-entropy Loss \cite{zhang2018crossentro}, \etc).

Unlike previous works that emphasize extracting individual-specific gait features, we provide a novel insight that a person's gait characteristics can also be described by the relationships between their own gait features and other gait anchors. What's more, we find that RD incorporating with GAs could further reduce the effect of covariates. 
%


\subsection{Open-set Recognition}
Open-set recognition (OSR) is a classification task with the additional requirement of rejecting input from unknown classes in the latent space \cite{openset}. 
The data labeled as ``unknown" classes should be far from the ``known'' training classes in the latent space. 
In the area of OSR, researchers mainly use the logits from the classifier to measure the distance between ``unknown" data and ``known" clusters.
OpenMax \cite{openset2} models all known classes by their logits as a single cluster and re-calibrates softmax scores according to the distance between input and other cluster centers. 
Several following works \cite{openset3,openset4,openset5,openset6} propose many mechanisms to improve the distance-based measures, such as data augmentation or introducing an ``other'' class. 
Recent works \cite{classifier, cen2023devil} find that a good close-set classifier can directly boost the performance of open-set recognition, leading to a reconsideration of classifiers.

Existing works in OSR also highlight the importance of classifiers, yet they primarily differ from our work in two key aspects: (1) The classifier in OSR is needed for ``known" class classification, while we drop the concept of classification and consider training identities as our GAs ; (2) OSR emphasizes the logits of unseen classes should be far from that of ``known" classes, while we advocate the discriminative capacity of logits. Overall, our work proposes a brand new perspective for the usage of classifiers.

\section{Our Approach}
In this work, we propose a new gait descriptor by revisiting the role of classifiers in the testing phase of gait recognition.  
We innovatively find the well-trained weights in the classifier can be regarded as gait anchors (GAs), and the relationship (logits) between the gait features and the set of gait anchors can be used as a discriminative descriptor. 
%

\subsection{Pipeline}
We follow a typical gait recognition procedure \cite{opengait, GaitGL} during the training phase. Given a training set with data-label pairs $\mathcal{D} = \{(x_i,y_i)\}$ where $x_i$ denotes a gait sequence and $y_i\in \{0,1, ... , C\}$ indicates the class label of $x_i$. 
%
%
The $\Omega$ is a feature extractor that embeds the gait sequences into a $d$-dimensional latent space, where $z_i = \Omega(x_i)$ is the extracted representation. 
The objective of training $\Omega$ on $\mathcal{D}$ is to acquire a discriminative transformation from $x_i$ to $z_i\in\mathbb{R}^d$, ensuring that the distance between $z$ from the same person is closer than those from different people in the latent space. 

\begin{figure}[t!]
 \centering
 \includegraphics[width=\linewidth]{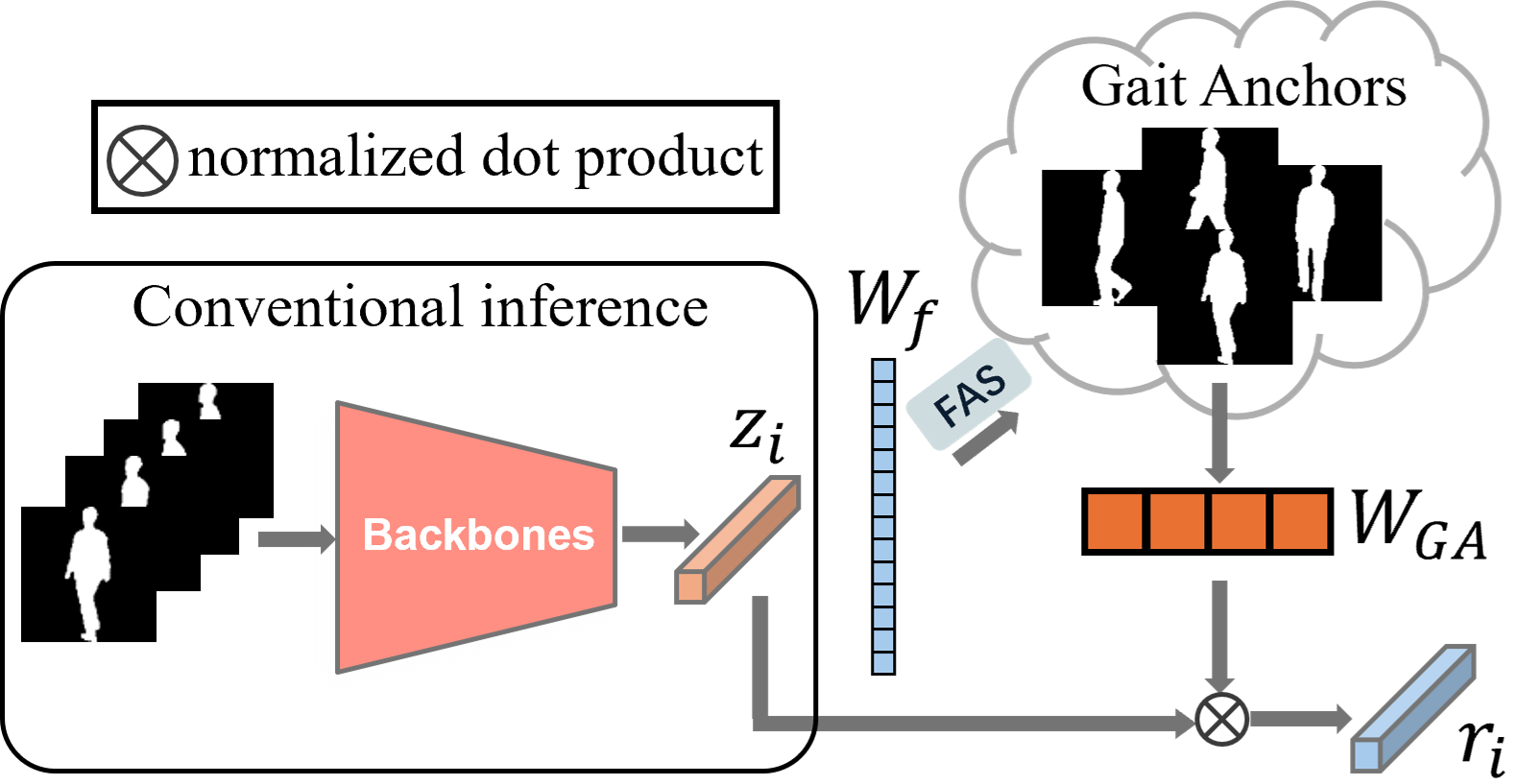}
 \caption{The overview of our pipeline. The gait anchors are selected from the well-trained classifier and the final representation is projected into a cosine similarity space.}
 \label{fig:pipeline}
\end{figure}

A common practice \cite{GLN,GaitGL,opengait,dou2023gaitgci,dou2022metagait} uses a combination of triplet loss \cite{triplet} and cross-entropy (CE) loss \cite{zhang2018crossentro} to get a discriminative $\Omega$.
A BNNeck classifier \cite{bnneck} $f:\mathbb{R}^{d} \rightarrow {\mathbb{R}^{C}}$ with a weights matrix $W_f\in\mathbb{R}^{d \times C}$ is introduced during the training stage for the classification task. 
In our work, we adopt a cosine similarity classifier \cite{cossimclass} in BNNeck. The $r_i=f(z_i)$ indicates the normalized dot product results of $z_i$ with $W_f$,  called logits. In this context, the logits can be seen as the descriptor of relationships,  where the larger $r_i^j$ is, the more similar $z_i$ and $W_f^j$ are.

Unlike prior works, we advocate the use $r_i$ for the final individual identification, thereby we keep the classifier in the test phase, as in \Cref{fig:pipeline}. 
To address the mentioned challenges brought by RD, we further adopt Farthest gait-Anchor Selection (FAS) algorithm to select the most discriminative set of GAs and Orthogonal Regularization Loss (ORL) for GAs overfitting problem.
Finally, with the help of the well-trained $W_f$ and our proposed methods, relation descriptors $r_i$ that achieve \textbf{higher performance} can be obtained \textbf{nearly without extra costs}, and their dimensions are \textbf{the same or even less} than that of $z_i$.

\subsection{A Novel Relation Descriptor}
We now introduce the core idea of our paper: gait goes beyond just an aggregation of individual features, and it can also be expressed through the relationships with the gait features of others.
In contrast to conventional methods that use the embedding $z_i$ as the final representation, we find that gait can be expressed by comparing the relationships between the gait features of different individuals, encompassing aspects of similarity, dissimilarity, common traits, \etc.
For example, given a random person's gait, it can be described as similar to \textit{gait anchor} 1 with $0.2$ cosine similarity, $0.7$ of GA2, $0.1$ of GA3, $-0.5$ of GA4, and so on. 
As a result, a descriptor of this person can be formulated as $r=[0.2, 0.7, 0.1, -0.5, ...]$, where each dimension of $r$ denotes the degree of similarity to a gait anchor.  
We term this novel descriptor \textbf{Relation Descriptor} (RD).
The Euclidean distance between two RDs can be defined as
\begin{equation}
    \label{eq:l2distR}
    dist(r_a,r_b) = \sqrt{\Sigma_i^{|GA|}(r_a^i-r_b^i)^2}
\end{equation}
where $|GA|$ is the number of pre-selected GAs. The potential discriminative capability of RD arises from the observation that gaits from the same person would share similar relationships to GAs, while gaits from different persons tend to have different ones.

Noting that the discriminative capability of the RD heavily relies on the set of GAs, a good GA should be relevant to ID information while remaining unbiased to covariates.
The higher the diversity of gait patterns within the GAs group is, the stronger the discrimination capability of the RD is.
To get a set of well-defined GAs, we set our sights on the classifier.
The weights $W_f$ in the classifier are trained by CE loss, resulting in each column vector in $W_{f,j}$ approximately equal to the semantic center of $j^{th}$ identity \cite{centerloss,zhang2018crossentro}. 
Hence, the well-trained $W_f$ in the classifier would encapsulate specific gait representations only tied to each identity in the training set.
Given this, the $W_f$ in the classifier is suitable for severing as GAs. 
Leveraging $W_f$, we can easily derive our new descriptor RD without any extra costs, formulated as
\begin{equation}
    r_i = \frac{W_f^T\cdot z_i}{||W_f||*||z_i||}
\end{equation}
where the $||\cdot||$ is the L2 norm. The normalized dot product between test gait features and $W_f$ represents similarity. And Eq. \ref{eq:l2distR} can be rewritten as
\begin{equation}
    d(r_a,r_b) = \sqrt{\Sigma_j^C(\frac{W_{f,j}^T}{||W_{f,j}^T||}\cdot(\frac{z_a}{||z_a||} -\frac{z_b}{||z_b||}))^2}
\end{equation}

It is easily noticed that $r_i$ is exactly the logits used by cosine CE loss \cite{cossimclass}, which is usually discarded in prior works since it represents the probability distribution of training classification that is inapplicable during testing. Nevertheless, through our novel perspective and experimental results, we argue that RD is more robust and generalizable. 

\noindent\textbf{Discussion: Why does RD work?} Gait is easily influenced by various covariates, including viewpoints, clothing changes, and more. 
Thus, a gait feature can be written as $z_i=\hat{z_i}+\epsilon_i$, where $\hat{z_i}$ is the invariant individual feature, and $\epsilon_i$ denotes the unexpected bias associated with covariates. 
%
An ideal gait recognition model should satisfy the conditions $\epsilon\rightarrow 0$ and $dist(z_a,z_p)=0<dist(z_a,z_n)-m$, where $a,p,n$ represents the anchor, positive and negative samples, respectively, indicating that $z$ is a consistent representation for the same person while maintaining distinctiveness across different individuals.
%
%
However, as discussed in \cite{wang2023causal}, it is hard to fulfill these two constraints simply by conventional metric learning, especially on real-world datasets.
%
The embedding $z_i$ inevitably incorporates bias $\epsilon_i$ to some extent. 

Inspired by image denoising \cite{denoise} and open-set recognition \cite{anchorclass}, projecting the original biased embedding $z_i$ onto several ID-relevant bases could suppress the ID-irrelevant bias $\epsilon_i$ and retain crucial ID-related information $\hat{z_i}$. 
This kind of projection decomposes $z_i$ into distinct semantic directions which can be measured by the cosine similarity between $z_i$ and GAs.
Fortunately, the well-trained $W_f$ converges toward different ID-related centers, expressed as $W_{f,j} \approx \frac{1}{|\mathcal{D}_j|}\sum_{z_i\in\mathcal{D}_j}(\hat{z_i}+\epsilon_i)$. Since samples of the $j^{th}$ person are usually collected across diverse covariant conditions, 
the term $\epsilon_i$ from different samples would be reduced by the average operator. Thus, the ID center would be less biased, formulated as $W_{f,j} \approx \hat{z_i} + \mathcal{O}(\epsilon)$, where $\mathcal{O}(\epsilon)$ is a minor term of $\epsilon$. Consequently, the $W_f$ is naturally more independent of covariates.

In summary, RD can be seen as the projection of $z$ onto different GAs, resulting in reduced susceptibility to the influence of covariates.

\subsection{Challenge: Dimension Expansion}
As we discussed above, the GAs can be seen as a set of semantic bases in the latent space, but, the number of bases relies on the number of training identities. 
When training on a large-scale dataset, the dimension of $W_f$ is inevitably expanding, incurring augmented storage costs and redundant information. 
Therefore, the combination of GAs should be carefully selected.

\noindent\textbf{What is a good combination of gait anchors?} The latent space is a manifold determined by the training data, and the class weights can be seen as bases that span the manifold. 
%
Intuitively, in order to maintain the discriminative ability, the manifold spanned by the combination of selected GAs needs to be as consistent as possible with the original manifold.
Additionally, when the number of identities is larger than the dimension of $z_i$, $C > d$, there are many linearly related bases in the classifier that can be removed.

Based on the above discussion, we assume that the most discriminative combination of GAs should be the one with the largest spanning space, \emph{i.e.}, the convex combinations of the GAs.

\noindent\textbf{How to select good gait anchors}.
Selecting the convex combinations of GAs from $W_f$ in the classifier is an NP-hard problem, which is generally difficult to solve in polynomial time. 
Hence, we propose a heuristic method, called Farthest gait-Anchor Selection (FAS), to solve this problem.

\begin{figure}[t!]
     \centering
     \includegraphics[width=\linewidth]{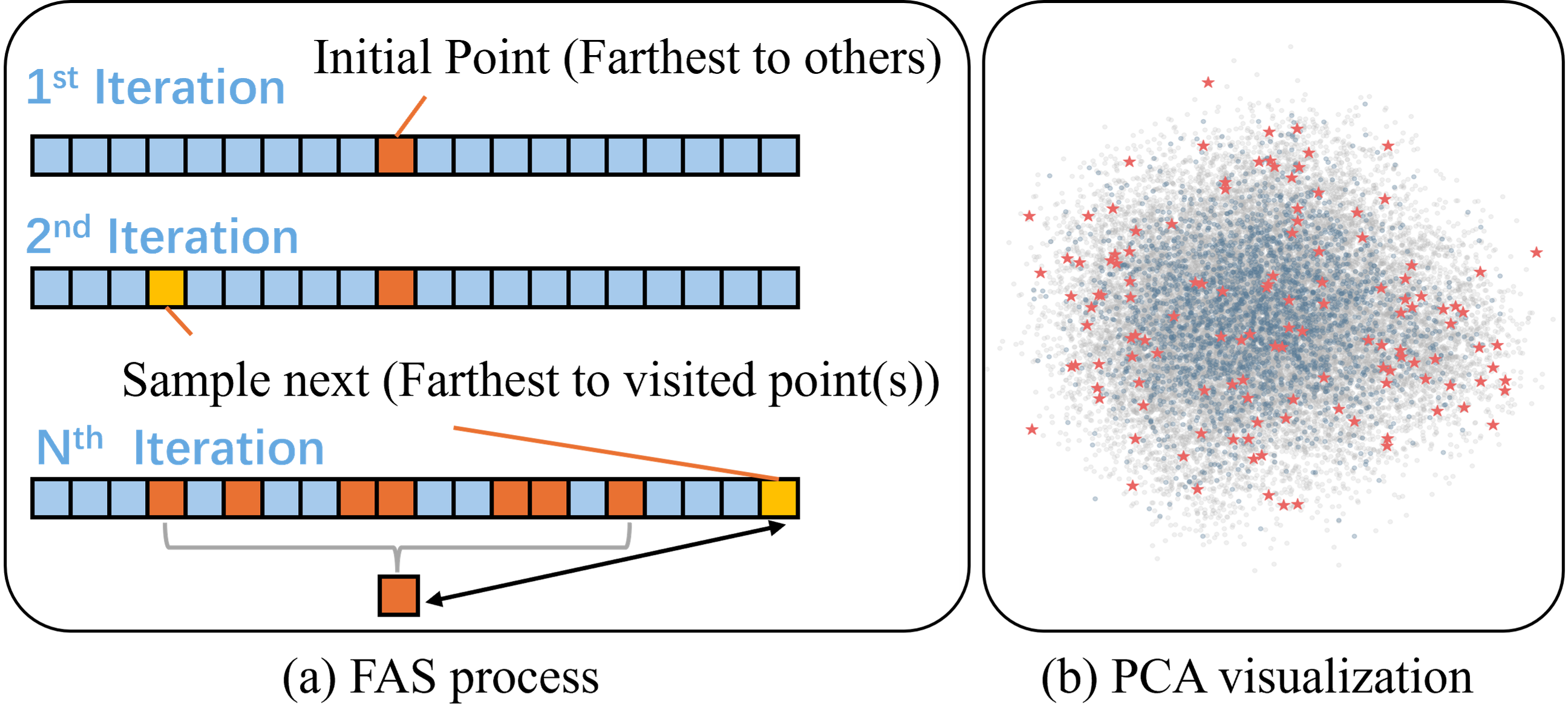}
     \caption{(a) Visualization of FAS process. (b) PCA visualization of training data (gray), $W_f$ (dark blue) in the classifier, and $W_{f,s}$ (red) selected by FAS. Zoom in for better clarity.}
     \label{fig:fas_tsne_class_weights}
 \end{figure}

Given weights in the classifier $\{W_{f,1}, W_{f,2}, ..., W_{f,C} \}$, we use iterative FAS to choose a combination of $N$ weights $\{W_{f,s1},W_{f,s2}, ..., W_{f,sN}\}$ as GAs, where the $W_{f,sj}$ is the farthest weights from the barycenter of the set $\{W_{f,s1},W_{f,s2}, ..., W_{f,sj-1}\}$, measured by Euclidean distances.
The final selected weights are denoted by $W_{f,s}\in\mathbb{R}^{d\times N}$.
%
%
FAS is a greedy algorithm described in Algorithm \ref{alg:farthest} and a visualization of this process is shown in \Cref{fig:fas_tsne_class_weights}.

The internal mechanism of FAS is based on the assumption that the farthest weight from the selected weights contains the most different semantic information, thus, it is good for increasing the diversity of the combination gait anchors. 
Compared to the original $W_f$, the $W_{f,s}$ collects the discriminative gait anchors and removes redundant ones.

\noindent\textbf{How to reduce the final dimension}. After filtering out some useless weights by FAS, the number of gait anchors is usually larger than the original embeddings' dimension. 
Luckily, we know the dimension of $W_{f,s}$ can be reduced to $d$ without information loss by Singular Value Decomposition (SVD) when $d<|GA|$. Note that directly applying SVD to reduce the dimension of $W_f$  cannot bring discriminative improvements in GAs, since SVD is an identical transformation of $W_f$. As a result, SVD here is only used to compress the knowledge in the $W_{f,s}$.

SVD is performed on the selected weight matrix $W_{f,s}$ to decompose the feature:
\begin{equation}
    W_{f,s} = U\Sigma V^T
\end{equation}
where $\Sigma\in\mathbb{R}^{d\times N}$ is rectangular diagonal singular value matrix, $U\in\mathbb{R}^{d\times d}$, and $V\in\mathbb{R}^{N\times N}$ are left and right orthogonal singular vector matrices, respectively. Then we select only top-$d$ weights from the $W_{f,s}$ by
\begin{equation}
    W_{GA} = U\Sigma[:,\,:\!d]
\end{equation}
where $W_{GA}\in\mathbb{R}^{d\times d}$ is the dimension-reduced GAs. 
With the selected GAs, we could calculate our final RD by $r=<z,W_{GA}>$, where $<\cdot, \cdot>$ denotes the cosine similarity.

\begin{algorithm}[t!]
    \caption{Pseudo-code of FAS in a PyTorch-like style.}
    \label{alg:farthest}
    \definecolor{codeblue}{rgb}{0.25,0.5,0.5}
    \lstset{
      backgroundcolor=\color{white},
      basicstyle=\fontsize{8pt}{8pt}\ttfamily\selectfont,
      columns=fullflexible,
      breaklines=true,
      captionpos=b,
      commentstyle=\fontsize{8pt}{8pt}\color{codeblue},
      keywordstyle=\fontsize{8pt}{8pt},
    }
    
    \begin{lstlisting}[language=python]
        # cdist(): matrix-wise L2 distance

        dist = cdist(W_f, W_f.t())
        # the farthest weight from all others
        farthest = dist.sum(-1).argmax()
        W_fs[0] = W[farthest]
        # remove the selected basis from W
        W_f.remove(farthest)
        
        for i in range(1, N):
            dist = cdist(W_fs.mean(dim=1), W.t())
            farthest = dist.sum(-1).argmax()
            W_fs[i] = W[farthest]
            W_f.remove(farthest)

    \end{lstlisting}
\end{algorithm}

\subsection{Challenge: GAs overfitting}
As previously discussed, the discrimination capability of RD relies on the diversity of gait patterns within the GA group, necessitating that each GA holds different semantic directions in the latent space. Consequently, the desirable GAs should share as little as correlated information, mathematically, requiring orthogonality in the latent space.

%
However, this requirement encounters challenges when the number of training identities is limited, particularly when $C < d$. In such instances, the classifier is prone to overfitting, leading to high inter-relatedness among GAs.


An analysis of the CE loss reveals a misalignment with the ideal characteristics of GAs.
During the training stage, the optimization target of CE is
\begin{equation}
    \mathcal{L}_{ce} = log(1+\sum_{y'\neq y_i}exp((r^{y'}-r^y)/T))
\end{equation}
where $T$ is the temperature and $r^y$ indicates the GT $y^{th}$ index of logits while $r^{y'}$ are the rest of logits. Notably, $L_{ce}$ encourages that $r^{y}$ is close to $1$ while $r^{y'}$ is close to $-1$. However, as previously mentioned, the desired $r^{y'}$ should be $0$ (orthogonal) instead of $-1$ (negative correlation), where $z$ only holds a positive correlation to GT class weight.

To tackle this limitation, we propose an orthogonal regularization loss (ORL) to minimize the correlation between different identity weights, increasing the diversity among GAs, which can be formulated as
\begin{equation}
    \mathcal{L}_{ORL} = \frac{1}{|\mathcal{B}|}\sum_{i=1}^\mathcal{B}(1-r_i^y + \frac{1}{C}\sum_{y'\neq y_i}|r^{y'}|)
\end{equation}
where $\mathcal{B}$ is the mini-batch, $C$ is the total number of training identities, and $|r^{y'}|$ denotes the absolute value of $r^{y'}$.

By enhancing the orthogonality between gait anchors and samples, the RD can better reflect the distinct characteristics of different individuals' gaits even with a small number of anchors.
A more carefully designed method is sure to further improve the performance, but it is not the priority of this paper.
Note that ORL is only used in small datasets with only a few identities.

\subsection{Optimization and Inference}
Our method can be built on top of off-the-shelf methods without extra parameters. The entire objective function can be formulated as
\begin{equation}
    \mathcal{L} = \mathcal{L}_{tri} + \mathcal{L}_{ce} + \lambda \mathcal{L}_{ORL}
\end{equation}
where the $\lambda$ is a hyper-parameter.

\section{Experiments}

\subsection{Datasets}
\noindent{\textbf{GREW}}\cite{GREW}.
GREW is the largest outdoor dataset, containing 26,345 subjects with 128,671 sequences captured from 882 cameras. According to its official partition, GREW is divided into three subsets, \ie, the training set with 20,000 subjects, the validation set with 345 subjects, and the test set with 6,000 subjects.

\noindent{\textbf{Gait3D}} \cite{Gait3d}.
Gait3D is an in-the-wild dataset, containing 4000 subjects with over 25,000 sequences captured from 39 cameras. Gait3D provides an official protocol that 3000 subjects are used for training while the remaining 1000 subjects are for test. 

\noindent{\textbf{OU-MVLP}} \cite{OUMVLP}.
The OU-MVLP is the largest indoor gait dataset under a fully controlled environment. It includes 10,307 subjects under normal walking conditions and 14 views. We adopt the widely-used protocol that 5153 subjects are used for training, and the rest are taken for the test. 

\noindent{\textbf{CASIA-B}} \cite{CASIAB}.
It is one of the most popular gait datasets, which contains 124 subjects from 11 view angles and 3 walking conditions: normal walking (NM), carrying bags (BG), and wearing a coat or jacket (CL). 
For fairness, our work follows the popular partition constructed by \cite{gaitset}. To be specific, the first 74 subjects are used for the training stage and the remaining 50 subjects are reserved for the test.

\noindent{\textbf{CCPG}} \cite{ccpg}. It is a clothing-changing dataset that provides 200 identities and over 16K sequences. Each identity has seven different cloth-changing statuses. CCPG-G is a subset of it, which provides off-the-shelf silhouettes for the gait recognition task. According to its official protocol, we use the first 100 identities for training and the rest for the test.
\begin{table}[t!]
    \centering
    \caption{Rank-1 accuracy (\%), Rank-5 accuracy (\%), Rank-10 accuracy (\%), and Rank-20 accuracy (\%) on GREW dataset.}
    \small
    \scalebox{0.88}{
    \begin{tabular}{c|cccc}
    \toprule
    Methods       & Rank-1 & Rank-5 & Rank-10 & Rank-20       \\ \midrule
    PoseGait \cite{posegait}      & 0.2                         & 1.0                         & 2.2                          & 4.3           \\ 
    GaitGraph \cite{teepe2021gaitgraph}    & 1.3                         & 3.5                         & 5.1                          & 7.5           \\ 
    GEINet \cite{geinet}       & 6.8                         & 13.4                        & 17.0                         & 21.0          \\ 
    TS-CNN \cite{ts-cnn}       & 13.6                        & 24.6                        & 30.2                         & 37.0          \\ 
    GaitSet \cite{gaitset}       & 46.3                        & 63.6                        & 70.3                         & 76.8          \\ 
    GaitPart \cite{gaitpart}     & 44.0                        & 60.7                        & 67.3                         & 73.5          \\ 
    GaitGL \cite{GaitGL}       & 47.3                        & 63.6                        & 69.3                         & 74.2          \\ 
    MGN \cite{MGN}          & 44.5                        & 61.3                        & 67.7                         & 72.7          \\ 
    CSTL \cite{cstl}         & 50.6                        & 65.9                        & 71.9                         & 76.9          \\ 
    MTSGait \cite{mtsgait}      & 55.3                        & 71.3                        & 76.9                         & 81.6          \\ \midrule
    GaitBase \cite{opengait}      & 60.1                        & 75.7                        & 80.5                         & 84.4          \\
    
    \rowcolor{gray!20}$\hookrightarrow$ \textit{\textbf{w.}} ours & \textbf{65.5}               & \textbf{78.7}               & \textbf{83.3}                & \textbf{86.3} \\ 
    \bottomrule
    \end{tabular}
    }
    \label{Tab:GREW}
\end{table}

\subsection{Implementation Details}
We utilize GaitBase \cite{opengait} as our main baseline on all datasets. The new version of GaitGL \cite{gaitGLTPAMI}, with a backbone of (64,128,128) channels, is also used on CASIA-B and CCPG for a comprehensive comparison. 
%
%
Moreover, we also reproduce GaitPart \cite{gaitpart} and GaitSet \cite{gaitset} in our ablation study. 
For these reproduced models \cite{gaitpart, gaitset}, we add an extra BNNeck cosine similarity classifier for them, and the temperature in CE loss is set to 16. 
For all methods, we follow their official training settings reproduced by OpenGait \cite{opengait}.

\noindent\textbf{Hyper-parameters}. The number of $N$ in FAS is set to 1024, 2048, and 8192 in the Gait3D, OU-MVLP, and GREW, respectively, and utilizing SVD to reduce the dimension of GAs to $d$ to ensure a fair comparison with the corresponding embedding. ORL is only used on CASIA-B and CCPG with $\lambda=1$ and $\lambda=0.1$, respectively.



\begin{table}[t!]
    \centering
    \caption{Rank-1 accuracy(\%), mAP(\%), and mINP(\%) comparison on Gait3D. The \textbf{bold} number denotes the best performances.}
    \small
    \scalebox{0.88}{
    \begin{tabular}{c|c|ccc}
    \toprule
    \multirow{2}{*}{Methods} & \multirow{2}{*}{Venue} & \multicolumn{3}{c}{Gait3D}                             \\ \cmidrule{3-5} 
                            &                            & Rank-1        & mAP           & mINP                  \\ \midrule
    PoseGait                & PR20                       & 0.2           & 0.5           & 0.3                    \\
    GaitGraph               & ICIP21                     & 6.3           & 5.2           & 2.4                     \\ \midrule
    GaitSet                 & AAAI19                     & 36.7          & 30.0          & 17.3                   \\
    GaitPart                & CVPR20                     & 28.2          & 21.6          & 12.4                  \\
    GLN \cite{GLN}                    & ECCV20                     & 31.4          & 24.7          & 13.6                    \\
    GaitGL                  & ICCV21                     & 29.7          & 22.3          & 13.3                   \\
    CSTL                    & ICCV21                     & 11.7          & 5.6           & 2.6                     \\
    SMPLGait \cite{Gait3d}               & CVPR22                     & 46.3          & 37.2          & 22.2                   \\ \midrule
    GaitBase                & CVPR23                     & 64.6          & 55.2          & 30.4                   \\
    \rowcolor{gray!20}$\hookrightarrow$ \textit{\textbf{w.}} ours  & -                          & \textbf{70.1} & \textbf{61.9} & \textbf{36.2}          \\ 
    \bottomrule
    \end{tabular}
    }
    \label{Tab:Gait3d}
\end{table}

\begin{table*}[ht!]
    \centering
    \caption{Averaged rank-1 accuracy (\%) on OU-MVLP, excluding identical-view cases.}
    \small
    \scalebox{0.88}{
    \begin{tabular}{c|cccccccccccccc|c}
    \toprule
    \multirow{2}{*}{Methods} & \multicolumn{14}{c|}{Prove View}                                                                                                                                                                                          & \multirow{2}{*}{Mean} \\ \cmidrule{2-15}
    \multicolumn{1}{c|}{}                         & 0$^\circ$             & 15$^\circ$            & 30$^\circ$            & 45$^\circ$            & 60$^\circ$            & 75$^\circ$            & 90$^\circ$            & 180$^\circ$           & 195$^\circ$           & 210$^\circ$           & 225$^\circ$           & 240$^\circ$           & 225$^\circ$           & 270$^\circ$           &                       \\ \midrule
    GEINet                                        & 23.2          & 38.1          & 48.0          & 51.8          & 47.5          & 48.1          & 43.8          & 27.3          & 37.9          & 46.8          & 49.9          & 45.9          & 45.7          & 41.0          & 42.5                  \\
    GaitSet                                       & 79.3          & 87.9          & 80.0          & 90.1          & 88.0          & 88.7          & 87.7          & 81.8          & 86.5          & 89.0          & 89.2          & 87.2          & 87.6          & 86.2          & 87.1                  \\
    GaitPrart                                     & 82.6          & 88.9          & 90.8          & 91.0          & 89.7          & 89.9          & 89.5          & 85.2          & 88.1          & 90.0          & 90.1          & 89.0          & 89.1          & 88.2          & 88.7                  \\
    GLN                                           & 83.8          & 90.0          & 91.0          & 91.2          & 90.3          & 90.0          & 89.4          & 85.3          & 89.1          & 90.5          & 90.6          & 89.6          & 89.3          & 88.5          & 89.2                  \\
    GaitGL                                        & 84.9          & 90.2          & 91.1          & 91.5          & 91.1          & 90.8          & 90.3          & 88.5          & 88.6          & 90.3          & 90.4          & 89.6          & 89.5          & 88.8          & 89.7                  \\ \midrule
    GaitBase                                      & 87.8          & 91.4          & 91.5          & 91.8          & 91.5          & 91.3          & 91.0          & 89.3          & 90.7          & 91.2          & 91.4          & 90.9          & 90.7          & 90.3          & 90.8
    \\
    \rowcolor{gray!20}$\hookrightarrow$ \textit{\textbf{w.}} ours                        & \textbf{88.8} & \textbf{92.0} & \textbf{92.0} & \textbf{92.2} & \textbf{91.9} & \textbf{91.6} & \textbf{91.4} & \textbf{90.3} & \textbf{91.3} & \textbf{91.4} & \textbf{91.6} & \textbf{91.3} & \textbf{91.0} & \textbf{90.8} & \textbf{91.3}         \\ 
    \bottomrule
    \end{tabular}
    }
    \label{Tab:OU-MVLP}
    \vspace{-0.2cm}
\end{table*}

\subsection{Performance Comparison}
\noindent\textbf{GREW}. We compare the performance of the proposed method with several gait recognition methods on GREW dataset and show experimental results in \Cref{Tab:GREW}. 
GREW is collected under an unconstrained condition, and it contains lots of unpredictable external covariates, such as occlusion and bad segmentation. 
As a result, gait sequences in the test set may be encoded by some unseen covariates that further produce meaningless gait representations.
From \Cref{Tab:GREW}, we can see the gait recognition methods that perform well on indoor datasets meet a large performance degradation. 
It shows the gait representations encoded by only individual features are not robust enough.
By replacing our relation descriptor incorporating FAS and SVD, our method elevates the accuracy of the state-of-the-art GaitBase by 5.4\% on Rank-1 accuracy. 
It is worth noting that our method adds no extra parameters and the dimension of the final gait representation is the same as the original GaitBase.
The experimental results indicate that the relation descriptor is more discriminative and robust in real-world scenarios.

\noindent\textbf{Gait3D}.Gait3D is also an unconstrained dataset. 
%
%
The comparison of prevailing competing methods is illustrated in \Cref{Tab:Gait3d}, demonstrating that our proposed method exhibits superior performance compared to previous methods by a considerable margin.
Our method measures the relationship between test gait and well-defined gait anchors, which is less affected by unseen covariates. 
As a result, our method outperforms all prevailing methods and boosts the performance of GaitBase by 5.4\%, 6.7\%, and 5.8\% on Rank-1, mAP, and mINP, respectively.

\noindent\textbf{OU-MVLP}. Since OU-MVLP is collected in a fully-constrained laboratory environment, its training and testing sets possess nearly identical covariates. 
The effects of covariates can be well eliminated during the training stage. 
Hence, directly using individual gait features as representation can achieve promising accuracy.
However,  as shown in \Cref{Tab:OU-MVLP}, we find our method can still boost the baseline performance on this dataset.
Our method boosts GaitBase accuracy across all viewpoints by using RD with the same dimension as the corresponding embedding.

\noindent\textbf{CASIA-B and CCPG-G}. CASIA-B and CCPG-G are also collected in a fully-constrained environment, containing viewpoints and clothing changes. It is worth noting that there are only 74 and 100 identities in the training set in CASIA-B and CCPG-G, which means the initial dimensions of RD are 74 and 100 on these two datasets, respectively. 
As shown in \Cref{Tab:CASIA-B_CCPG}, even though the dimension of RD is fewer than the original embeddings, our method still achieves higher accuracy.
Our method improves the average mAP of Gaitbase by 2.7\% on CCPG and 0.6\% on CASIA-B. 
Since GaitGL performs better than GaitBase on CASIA-B, we also adapt our method to GaitGL, and the results on both datasets again verify the effectiveness of our method. 
The experiments have demonstrated that relation descriptors exhibit discriminative capability comparable to or higher than individual gait features.


\begin{table}[t!]
    \centering
    \caption{Averaged rank-1 accuracy (\%) on CASIA-B and rank-1 accuracy (\%) and mAP on CCPG-G. The `*' denotes that the results are based on our strict reproduction by OpenGait.}
    \small
\scalebox{0.88}{
\begin{tabular}{c|ccc|ccc}
\toprule
\multirow{2}{*}{Method} & \multicolumn{3}{c|}{CASIA-B} & \multicolumn{3}{c}{CCPG-G (Rank-1 $|$ mAP)}                       \\ \cmidrule{2-7} 
                        & NM       & BG      & CL      & CL                  & UP                  & DN                  \\ \midrule
GaitSet                 & 95.8     & 90.0    & 75.4    & 77.7$|$46.5          & 83.5$|$59.6          & 83.2$|$60.1          \\
GaitPart                & 96.1     & 90.7    & 78.7    & 77.8$|$45.5          & 84.5$|$63.1          & 83.3$|$60.1          \\
CSTL                    & 98.0     & \textbf{95.4}    & 87.0    & -                   & -                   & -                   \\ \midrule
GaitGL*                 & 97.7     & 94.7    & 86.0    & 81.9$|$50.5          & 91.2$|$71.0          & 86.4$|$67.0          \\
\rowcolor{gray!20}$\hookrightarrow$ \textit{\textbf{w.}} ours                   & 97.5     & 95.0    & \textbf{87.8}    & 82.4$|$51.8 & 91.4$|$72.3 & 87.2$|$69.5 \\ \midrule
GaitBase*               & 97.6     & 94.0    & 77.4    & 91.8$|$64.6          & 95.3$|$78.3          & 94.7$|$79.5          \\
\rowcolor{gray!20}$\hookrightarrow$ \textit{\textbf{w.}} ours                   & \textbf{98.1}     & 94.1    & 77.9    & \textbf{92.3$|$67.3} & \textbf{95.5$|$81.6} & \textbf{95.3$|$81.6} \\ 
\bottomrule
\end{tabular}
}
\label{Tab:CASIA-B_CCPG}
\end{table}

\begin{table}[t!]
    \centering
    \caption{Ablation study on relation descriptors (RD), Farthest gait-Anchor Selection (FAS), and Orthogonal Regularization Loss (ORL) on CASIA-B with GaitGL and Gait3D with GaitBase.}
    \small
    \scalebox{0.88}{
    \begin{tabular}{c|ccc|ll}
    \toprule
    \multirow{2}{*}{} & \multirow{2}{*}{RD} & \multirow{2}{*}{FAS} & \multirow{2}{*}{ORL} & \multicolumn{1}{c}{CASIA-B}   & \multicolumn{1}{c}{Gait3D}      \\ \cmidrule{5-6} 
                      &                     &                      &                      & Mean Acc. $\uparrow$ & Rank-1 Acc. $\uparrow$ \\ \midrule
    \#1               &                     &                      &                      & 92.8         & 64.6        \\
    \#2               & \checkmark           &                      &                      & 91.4\textcolor{red}{$^{\downarrow-1.4}$}         & 68.8\textcolor{ao}{$^{\uparrow+2.2}$}        \\
    \#3               & \checkmark           & \checkmark            &                      & \textit{N/A}         & \textbf{70.1}\textcolor{ao}{$^{\uparrow+5.5}$}        \\
    \#4               & \checkmark           &                      & \checkmark            & \textbf{93.4}\textcolor{ao}{$^{\uparrow+0.6}$}         & \textit{N/A}          \\ 
    \bottomrule
    \end{tabular}
    }
    \label{Tab:Abl}
\end{table}

\begin{table}[t!]
    \centering
    \caption{Model-agnostic results on Gait3D dataset.}
    \small
    \scalebox{0.88}{
    \begin{tabular}{c|llll}
    \toprule
    Rank-1          & \multicolumn{4}{c}{Methods}                                                                                            \\ \cmidrule{2-5} 
    Acc.                                                                              & \multicolumn{1}{c}{GaitSet} & \multicolumn{1}{c}{GaitPart} & \multicolumn{1}{c}{GaitGL} & \multicolumn{1}{c}{GaitBase} \\ \midrule
    Baseline                                                                     & 36.7                        & 28.2                         & 29.7                       & 64.6                         \\
    \rowcolor{gray!20}$\hookrightarrow$ \textit{\textbf{w.}} ours                                                                         & 42.5\textcolor{ao}{$^{\uparrow+5.8}$}                           & 38.0\textcolor{ao}{$^{\uparrow+9.8}$}                            & 37.1\textcolor{ao}{$^{\uparrow+7.4}$}                         & 70.1\textcolor{ao}{$^{\uparrow+5.5}$}                         \\ 
    \bottomrule
    \end{tabular}
    }
    \label{Tab:model-agno}
    \vspace{-0.2cm}
\end{table}

\begin{figure}
     \centering
     \includegraphics[width=0.7\linewidth]{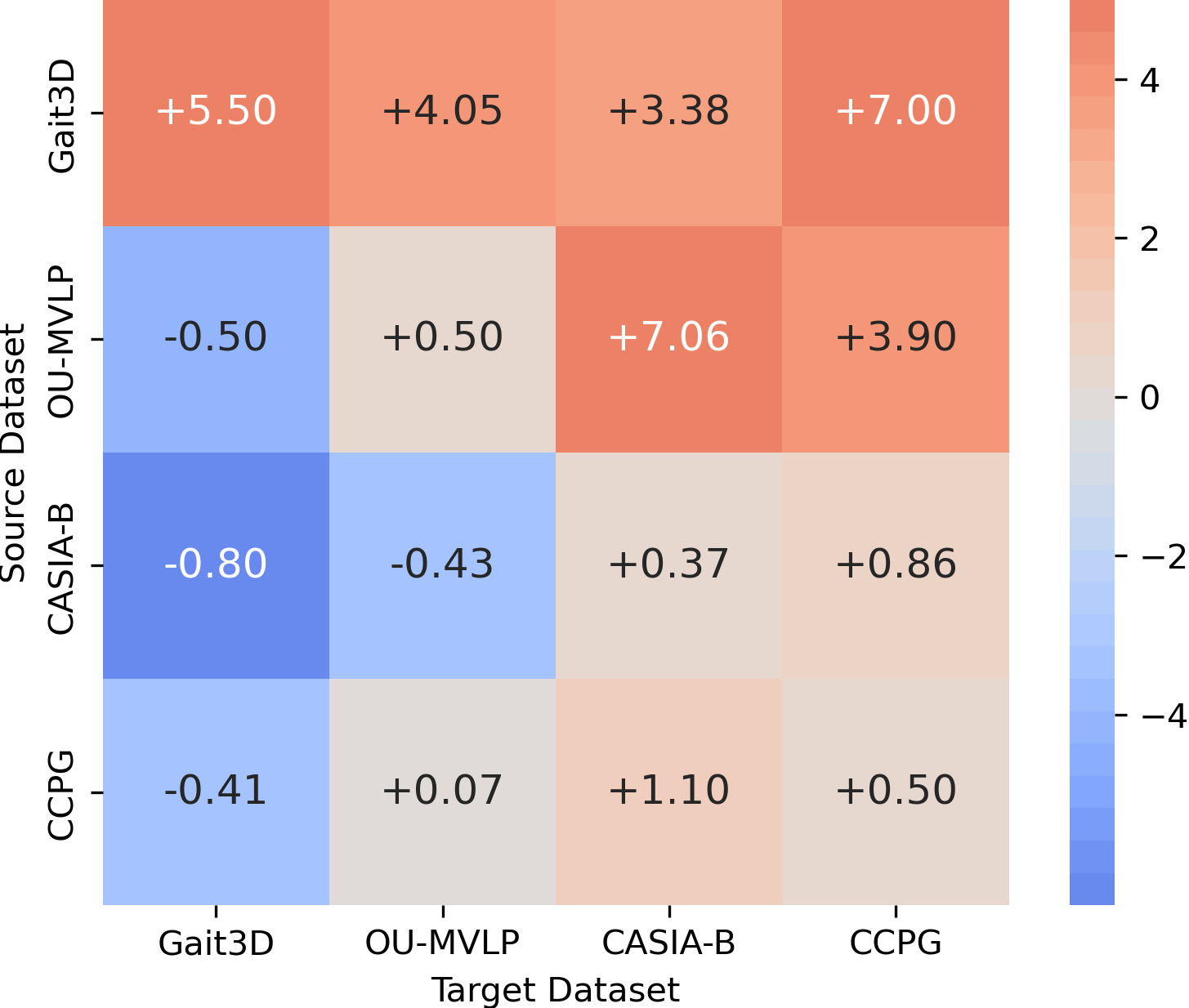}
     \captionof{figure}{The gain of recognition accuracy ($+/-$) brought by RD $r$ for cross-domain testing compared to using embedding $z$ among Gait3D, OU-MVLP, CASIA-B, and CCPG.}
     \label{fig:cross-domain} 
\end{figure}
\subsection{Ablation Study}
In this subsection, we provide the ablation study of each component in our method on CASIA-B and Gait3D.

\noindent\textbf{Analysis of RD, FAS, and ORL.}
The ablation results are illustrated in \Cref{Tab:Abl}. 
Note that FAS is only employed for datasets where the number of identities $C$ in the training set exceeds the output dimension $d$. 
In contrast, ORL is only applied on datasets where $C<d$.
Here is the analysis: 1) from experiment \#2, RD improves the baseline by 2.2\% on Gait3D but degrades the performance on CASIA-B by 1.4\% due to fewer gait anchors;
2) Comparing \#4 with \#2, adding ORL to generate more separated GAs brings improvement on CASIA-B, which shows the effectiveness of ORL;
3) Comparing \#4 with \#2, the selected combination of gait anchors by FAS can further improve the recognition accuracy.
%
Overall, the ablation result verifies the effectiveness of our proposed methods.

\begin{figure}[ht!]
    \centering
     \begin{subfigure}[b]{0.48\linewidth}
         \centering
         \includegraphics[width=\linewidth]{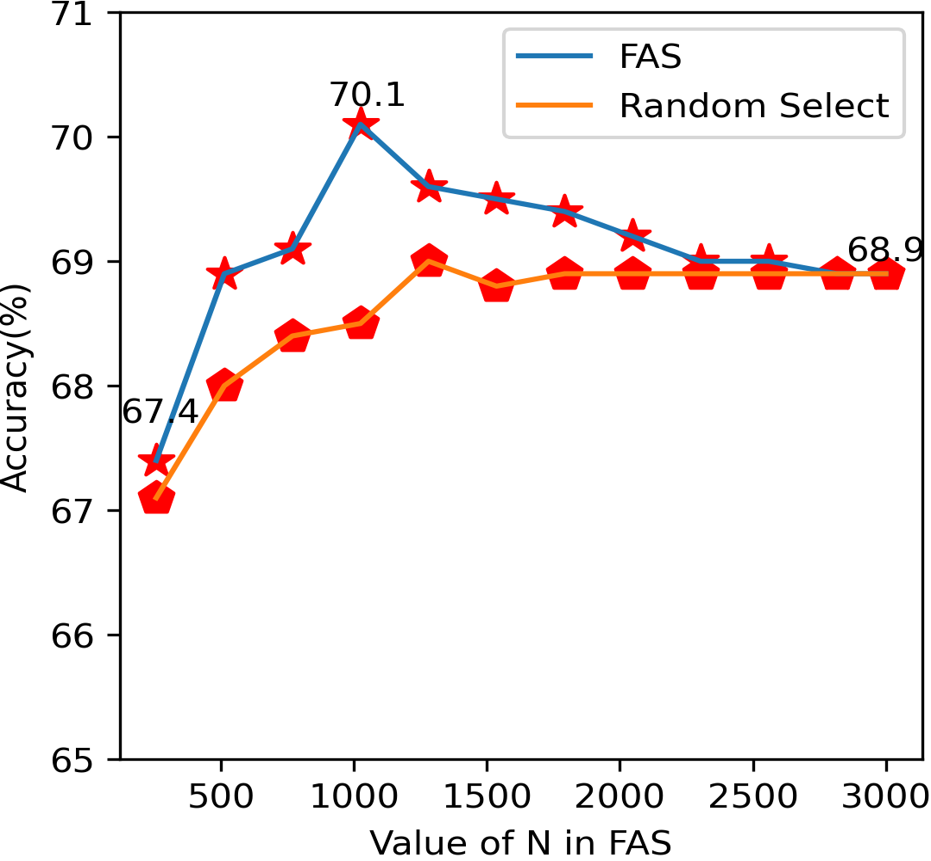}
         \caption{ }
         \label{fig:N_value}
     \end{subfigure}
     \begin{subfigure}[b]{0.48\linewidth}
         \centering
         \includegraphics[width=\linewidth]{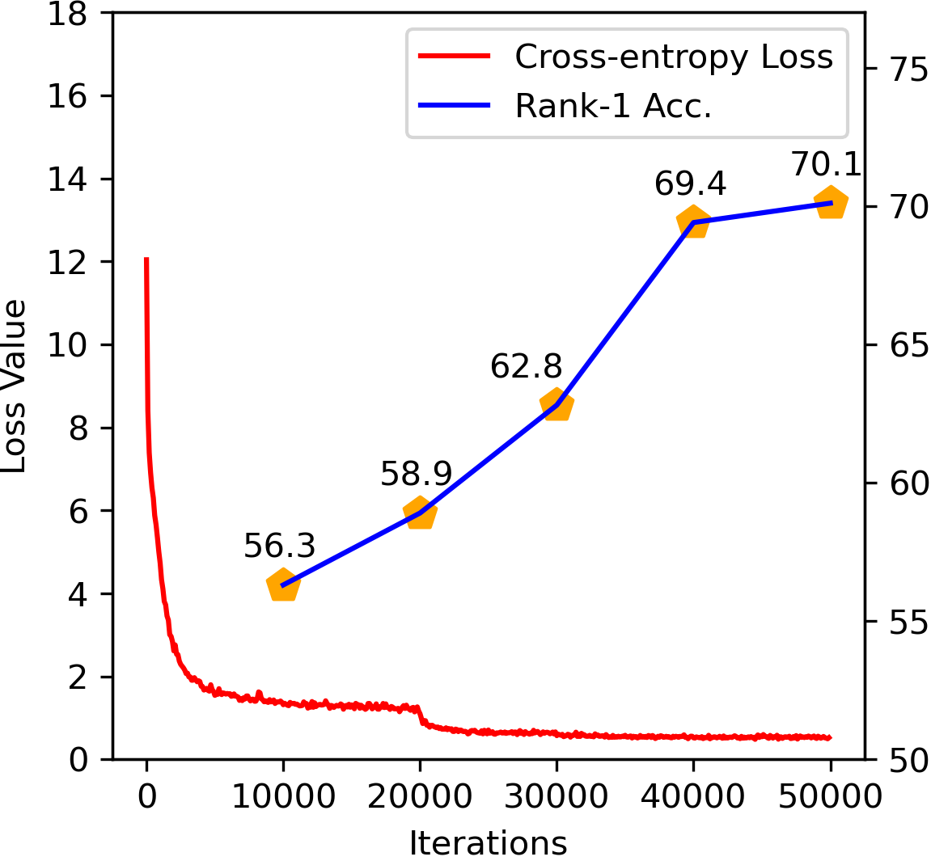}
         \caption{ }
         \label{fig:loss_acc_curve}
     \end{subfigure}
     \caption{(a) Ablation study on the number of selected gait anchors in FAS on Gait3D. (b) The cross-entropy loss and accuracy curves of GaitBase on Gait3D.}
     \vspace{-0.5cm}
\end{figure}

\noindent\textbf{Analysis of model-agnostic results.}
As shown in \Cref{Tab:model-agno}, we conduct experiments on Gait3D with four popular methods, including GaitSet, GaitPart, GaitGL, and GaitBase.
Since GaitSet and GaitPart don't use CE loss, we train an extra cosine similarity classifier for them.
%
%
The results show that the proposed method can improve performance regardless of the baseline backbones or network structures. 
The model-agnostic property further verifies the superior discriminative capacity of the relation descriptor.

\noindent\textbf{Analysis of the generalizable ability on cross-domain testing}. As shown in \Cref{fig:cross-domain}, we compare the cross-domain recognition performance between RD $r_i$ and embedding $z_i$ of GaitBase. In most cases, RD gains a better rank-1 accuracy, such as 7.0\% improvements on Gait3D to CCPG compared to embedding. This experiment demonstrates that the projection-based RD can potentially alleviate the domain gap.

\noindent\textbf{Analysis of the value $N$ in FAS.}
As mentioned above, not all weights in the classifier contribute to better recognition performance. 
\Cref{fig:N_value} shows that the performance of relation descriptors relies on the pre-selected gait anchors. 
Moreover, it is important to note that selecting too few or all of the weights doesn't necessarily lead to performance improvement, where the curve first rises and then falls as $N$ is increased.
Compared to random selection, FAS can bring consistent performance improvement.

\noindent\textbf{Analysis of the relationship between classifier convergence and accuracy.} The relation descriptor depends on the well-defined weights in the classifier, which is based on the assumption that each weight represents a typical gait representation. 
The CE loss incorporating with ORL requires separating the weights of different identities. Hence, the weights in a classifier with better convergence could represent more distinct gait features, leading to superior results. 
This phenomenon can be observed in \Cref{fig:loss_acc_curve}.

\begin{figure}[t!]
     \centering
     \begin{subfigure}[b]{0.46\linewidth}
         \centering
         \includegraphics[width=\linewidth]{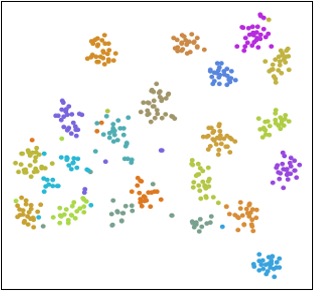}
         \caption{Embedding}
         \label{fig:tsne_baseine}
     \end{subfigure}
     \begin{subfigure}[b]{0.46\linewidth}
         \centering
         \includegraphics[width=\linewidth]{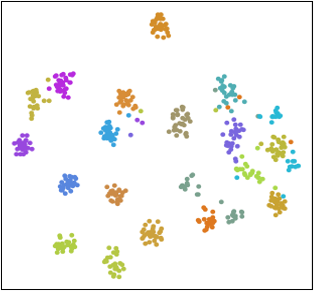}
         \caption{RD}
         \label{fig:tsne_rd}
     \end{subfigure}
     
        \caption{T-SNE visualization on Gait3D test set with randomly selected 20 classes. One color denotes a class.}
        \label{fig:TSNE}
\vspace{-0.5cm}
\end{figure}

\noindent\textbf{Visualization}
We visualize the feature distribution of models' original embeddings and RD on Gait3D test set with 20 randomly selected classes. 
By comparing \Cref{fig:tsne_baseine,fig:tsne_rd}, it shows that the intra-class variation is further reduced and the inter-class distance is hence enlarged by our relation descriptor. 
The visualization demonstrates that RD is a more discriminative representation.

\section{Limitations and Conclusion}
For limitations, GA relies on the label of identities in the training set, which limits its application to unsupervised learning. Besides, when covariates are effectively diminished during training, \eg, OUMVLP, RD and embedding share a comparable performance. Future efforts may involve integrating basis generation techniques to address the first issue and fine-grained attribute-based methods to alleviate the latter.

In conclusion, we provide a new perspective that gait can be expressed by a relation descriptor by projecting to ID-related GAs, which offers a way to diminish the bias in original embeddings. Further, we revisit the role of the well-trained weights in the classifier, arguing that they are exactly as suitable GAs.
Based on this finding, we propose a novel relation descriptor serving as a more discriminative representation of gait recognition. 
Besides, FAS and ORL are carefully designed to solve dimensionality challenges brought by RD and further boost recognition performance.
Overall, we hope these new insights could prompt further research in the gait community.

{
    \small
    \bibliographystyle{ieeenat_fullname}
    \bibliography{main}
}


\end{document}